\pdfoutput=1

\documentclass[11pt]{article}

\usepackage{ACL2023}

\usepackage{times}
\usepackage{latexsym}
\usepackage{todonotes}
\usepackage[T1]{fontenc}
\usepackage{multirow}
\usepackage{booktabs}
\usepackage{tikz-dependency}
\usepackage{amsmath}
\usepackage[normalem]{ulem}
\usepackage{lscape}

\usepackage[utf8]{inputenc}

\usepackage{microtype}

\usepackage{pifont}
\newcommand{\cmark}{\ding{52}}%
\newcommand{\xmark}{\ding{56}}

\title{RED\textsuperscript{FM}: a Filtered and Multilingual Relation Extraction Dataset}

\author{
Pere-Llu\'is Huguet Cabot\\
  Babelscape, Italy \\
  \& Sapienza University of Rome\\
\texttt{huguetcabot@babelscape.com} \\\And
    Simone Tedeschi\\
  Babelscape, Italy \\
  \& Sapienza University of Rome\\\texttt{tedeschi@babelscape.com} \\\AND
  Axel-Cyrille Ngonga Ngomo\\
  Paderborn University  \\
  \texttt{axel.ngonga@upb.de} \\ \And
  Roberto Navigli \\
  Sapienza University of Rome \\
  \texttt{navigli@diag.uniroma1.it} \\}

\begin{document}
\maketitle
\begin{abstract}

Relation Extraction (RE) is a task that identifies relationships between entities in a text, enabling the acquisition of relational facts and bridging the gap between natural language and structured knowledge. However, current RE models often rely on small datasets with low coverage of relation types, particularly when working with languages other than English. \\
In this paper, we address the above issue and provide two new resources that enable the training and evaluation of multilingual RE systems.
First, we present SRED\textsuperscript{FM}, an automatically annotated dataset covering 18 languages, 400 relation types, 13 entity types, totaling more than 40 million triplet instances. Second, we propose RED\textsuperscript{FM}, a smaller, human-revised dataset for seven languages that allows for the evaluation of multilingual RE systems. 
To demonstrate the utility of these novel datasets, we experiment with the first end-to-end multilingual RE model, mREBEL, 
that extracts triplets, including entity types, in multiple languages. We release our resources and model checkpoints at \href{https://www.github.com/babelscape/rebel}{https://www.github.com/babelscape/rebel}. 
\end{abstract}

\section{Introduction}
The vast majority of online and offline content consists of raw, natural language text containing factual information. 
Current Large Language Models (LLMs) are pretrained on such text, allowing reasoning over it through tasks such as Question Answering \cite{BOUZIANE2015366}
or Text Summarization \cite{el2021automatic}. On the other hand, structured resources such as Knowledge Graphs enable knowledge-based, explainable, machine-ready reasoning over their content. Both approaches are important and are widely used within Natural Language Processing systems, with recent trends looking at combining them \cite{yamada-etal-2020-luke, sun2021ernie}. 

Information Extraction tackles the need for systems that extract structured information from raw text. Specifically, end-to-end Relation Extraction extracts the relational information between entities in a given text, providing a structured prediction. However, although some highly capable systems have been released \cite{wang-lu-2020-two, tanl, huguet-cabot-navigli-2021-rebel-relation}, there are few high-quality, contemporary resources. Current RE datasets are outdated, behind paywalls, have design flaws, or only consider English. While multilingual datasets exist, such as ACE05\footnote{\url{https://catalog.ldc.upenn.edu/LDC2006T06}} or SMiLER \cite{seganti-etal-2021-multilingual}, the former covers only six relation types, and requires a paid license for its use. The latter is more recent, bigger, and has a higher coverage of relation types, but it does not contain human-annotated samples that permit reliable evaluation and is not conducive to train End-to-End Relation Extraction systems. Instead, the availability of large high-quality resources is fundamental in order to allow LLMs to be trained and evaluated on trustworthy multilingual RE benchmarks. 

In this paper, we introduce large amounts of high-coverage RE annotated data in a multilingual fashion. Our new resources will enable the training of multilingual RE systems and their evaluation. 
In particular, we provide three main contributions:

\begin{enumerate}
\item We present RED\textsuperscript{FM}, our humanly-revised dataset with 32 relation types and 7 languages. 
\item We introduce SRED\textsuperscript{FM}, a silver-standard dataset 
based on interconnecting Wikipedia and Wikidata, filtered by a Critic system trained on human annotations. It covers 400 relation types, 18 languages, and more than 44M triplet instances. Both datasets are automatically enriched with entity-type information using a novel entity typing approach. 
\item We demonstrate the usefulness of these new resources by 
releasing 
mREBEL, a multilingual system for Relation Classification and Relation Extraction that extracts entity types. 
\end{enumerate}

\section{Related work}
\subsection{Relation Extraction}

In Relation Extraction (RE), the goal is to identify all triplets, composed of a subject, an object, and a relation between them, within a given text. Early approaches to RE split the task into two different sub-tasks: Named Entity Recognition (NER) \cite{nadeau2007survey}, which identifies all entities, and Relation Classification \cite{bassignana-plank-2022-mean}, which  classifies the relationship, or lack thereof, between them. 
However, errors from the NER system may be propagated to the subsequent module, leaving the shared information in the interaction of both tasks unexplored.

Recent works have tackled RE in an end-to-end fashion, seeking to overcome these problems by using different abstractions of the task. \citet{miwa-sasaki-2014-modeling} introduced a table representation and reframed RE as a table-filling task. This idea was further explored and extended by \citet{pawar-etal-2017-end} and \citet{wang-lu-2020-two}. However, these systems still had some restrictions, such as assuming that only one relation exists between each pair. Instead by framing the task as a sequence of triplets to be decoded, seq2seq approaches \cite{tanl, huguet-cabot-navigli-2021-rebel-relation} provided more flexibility to the RE task and lifted some of these restrictions. Nevertheless, seq2seq models are notoriously data-hungry, hence vast amounts of data are needed to enable them to learn the task with satisfactory scores.

\subsection{Relation Extraction Datasets}
Manually annotating RE data is a costly and time-consuming process. As a result, many RE datasets have been created using distant supervision methods, such as NYT \cite{10.1007/978-3-642-15939-8_10}, T-REx \cite{elsahar-etal-2018-rex} or DocRED \cite{yao-etal-2019-docred}.
Despite their widespread use in the RE community, these datasets have limitations. For instance, automatically generated datasets often contain noisy labels, leading to unfair or misleading evaluations. Additionally, there has been a long-standing focus on monolingual relation extraction systems, particularly in English. 

The ACE05 benchmark presented some of the first relation extraction datasets in three languages, Arabic, Chinese, and English. However the focus on Arabic and Chinese quickly faded away while resources for English continued to grow.
One of the main challenges in developing multilingual relation extraction systems is the lack of annotated data for the task. 
The SMiLER dataset \cite{seganti-etal-2021-multilingual}, based on distant supervision, uses Wikipedia and Wikidata to create a multilingual relation extraction dataset. However, besides being automatic, SMiLER limits annotations to one triple per sentence. 
With this paper, we overcome the limitations of existing datasets by providing a new multilingual evaluation dataset that includes manual annotations and enables RE with a wide coverage and higher quality despite being based on automatic annotation.

\begin{figure*}[t!]
    \centering
    \def\svgwidth{2\columnwidth}
    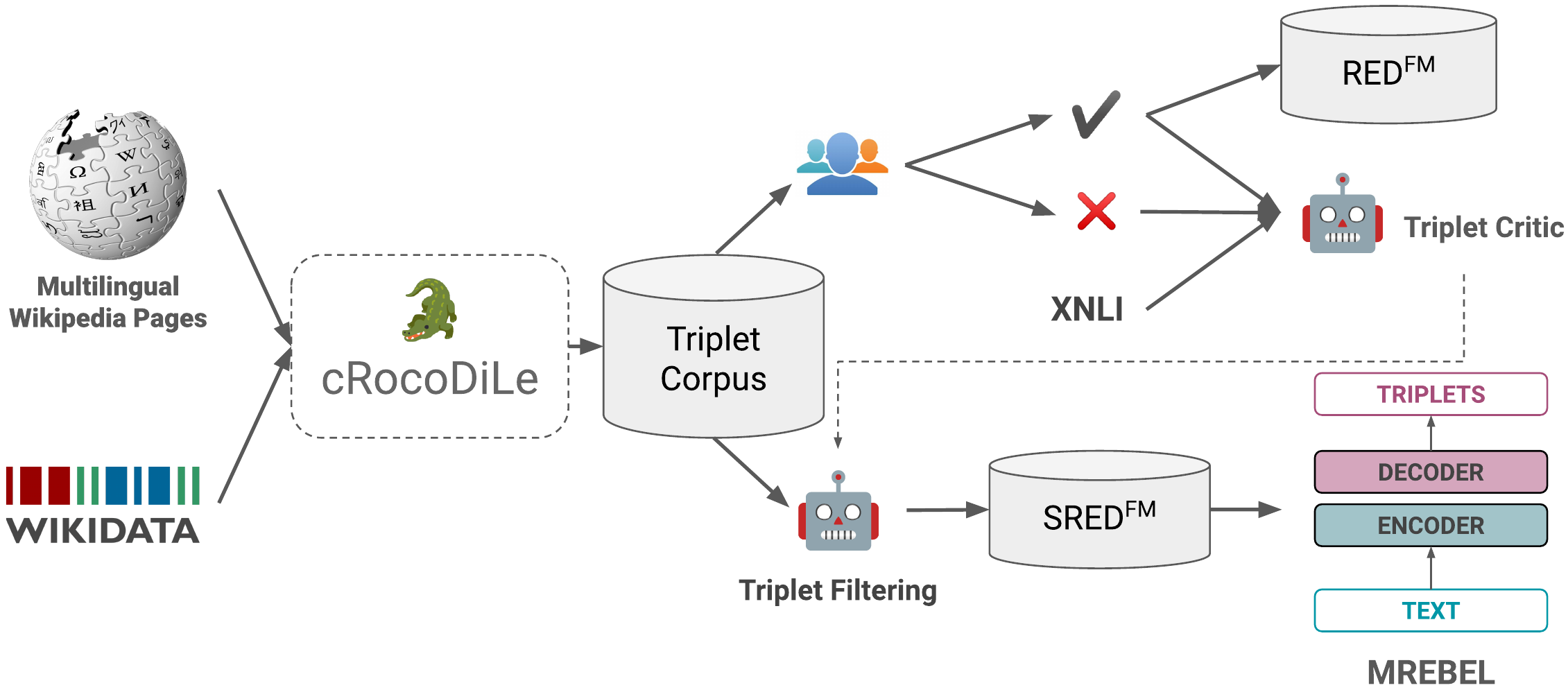
    \caption{Our full pipeline for the creation of RED\textsuperscript{FM}, SRED\textsuperscript{FM} and mREBEL.}
    \label{fig:dev_corr_guided}
\end{figure*}
\section{RED\textsuperscript{FM}}

In this Section, we present RED\textsuperscript{FM}, our supervised and multilingual dataset for Relation Extraction, and a larger SRED\textsuperscript{FM}, a silver-annotated dataset covering more languages and relation types. The creation of the dataset consists of several steps: data collection and processing (Section \ref{sec:data-extraction}), manual annotation (Section \ref{sec:annotation}), a triplet filtering system (Section \ref{sec:critic}) and entity typing (Section \ref{sec:entity-typing}). Figure \ref{fig:dev_corr_guided} shows an overview of this process.

\subsection{Data Extraction}\label{sec:data-extraction}
We base our dataset on Wikidata and Wikipedia, and expand cRocoDiLe, the data extraction pipeline from \citet{huguet-cabot-navigli-2021-rebel-relation}, to obtain a large collection of triplets in multiple languages (see Appendix \ref{sec:crocodile} for more details). We use the hyperlinks from Wikipedia abstracts, i.e. the content before the Table of Contents, as entity mentions and the relations in Wikidata between them. 
We run our pipeline in the following 18 languages: Arabic, Catalan, Chinese, Dutch, German, Greek, English, French, Hindi, Italian, Japanese, Korean, Polish, Portuguese, Russian, Spanish, Swedish, and Vietnamese. Then, we collapse inverse relations and 
keep the 400 most frequent ones.
We highlight that some extracted relations are not necessarily entailed by the Wikipedia text; therefore, we apply a multilingual NLI system\footnote{\href{https://huggingface.co/joeddav/xlm-roberta-large-xnli}{xlm-roberta-large-xnli}} to filter out those with a low entailment score (i.e. <\,0.1). 

Despite using NLI techniques to filter out false positives, distant RE annotations still present noisy labels. This can result in unfair or misleading evaluations, as demonstrated for TACRED \cite{zhang-etal-2017-position}, which had 23.9\% wrongly annotated triplets that were later revised and corrected by \citet{stoica2021retacred}. 
Moreover, our triplet corpus extraction pipeline relies on existing triplets in Wikidata, similar to T-REx \cite{elsahar-etal-2018-rex}. These latter showed how certain relation types, such as ``capital'', have a lower entailment score and may not be entailed by a given text even though the entities involved share the relation in Wikidata.

Given these challenges in distant RE annotation, manual filtering of a portion of the data is necessary to ensure high-quality, accurate annotations.
\subsection{Manual Annotation}\label{sec:annotation}
We manually filter a portion of the data to deal with false positives present in the dataset for a subset of languages (i.e. Arabic, Chinese, German, English, French, Italian, and Spanish) through crowdsourced annotation:
\begin{enumerate}
    \item We reduce the coverage of the annotated data to the top 32 most frequent relation types. See Appendix \ref{sec:appendix_annotation} for details on each of these types.
    \item We select a portion of our silver annotated data consisting of i) common Wikipedia pages across those languages and ii) a random sample with less frequent relations to balance the dataset.
    \item We ask human annotators to validate each triplet. They are shown the context text with subject and object entities highlighted, and the possible relation between them from the silver extraction. They must answer whether the text conveys the necessary information to infer that the relationship between those two entities is true.
    \item We annotate each triple three times using different annotators, obtaining an average inter-rater reliability (Krippendorff's alpha) across languages of $\alpha_\kappa$ = 0.73.
    \item We keep as true positives those relations with at least two annotators answering true. We consider the rest false positives.
\end{enumerate}
\begin{table}[t]
\resizebox{\columnwidth}{!}{
\begin{tabular}{lrrrrrrr}\toprule
                         & \textbf{ar} & \textbf{de} & \textbf{fr} & \textbf{en} & \textbf{es} & \textbf{it} & \textbf{zh} \\ \midrule
\textbf{Annotators}      & 3           & 9           & 12          & 9           & 10          & 13          & 7           \\
\textbf{$\alpha_\kappa$} & 0.76        & 0.80        & 0.77        & 0.72        & 0.61        & 0.73        & 0.70   \\ 
\textbf{Filtered (\%)}        & 6.79      & 5.47     & 4.40     & 8.54      & 12.53     & 8.93      & 12.11    \\
\bottomrule    
\end{tabular}}
\caption{Number of annotators, their agreement based on Krippendorff's alpha, and the percentage of filtered triplets for each language.}
\label{tab:annotation}
\end{table}

We employed Amazon Mechanical Turk and manually selected annotators that qualified for the task in each language. The annotation scheme can be found in Appendix \ref{sec:appendix_annotation}. From Table \ref{tab:annotation} we see that around 8\% of annotated triplets, on average, were labeled as non-entailed by the context provided, albeit the percentage varies across languages. For instance, Spanish had a lower agreement across annotators and a higher number of filtered instances.

\begin{table}[t]
\resizebox{\columnwidth}{!}{
\begin{tabular}{cl|cc|ccc} \toprule\toprule
                                & \multicolumn{1}{c}{} & \multicolumn{1}{c}{\textbf{All True}} & \multicolumn{1}{c|}{\textbf{XNLI}} & \multicolumn{1}{c}{\textbf{Ours}} & \multicolumn{1}{c}{\textbf{+XNLI$^{-}$}}  & \multicolumn{1}{c}{\textbf{+XNLI}} \\ \midrule
\multirow{4}{*}{\rotatebox[origin=c]{90}{\textbf{Arabic}}}    & \textbf{R.}      & 100.0                               & 76.6                                 & 96.9                               & \textbf{98.6}                                                                          & 98.5                            \\
                                & \textbf{P.}   & ~~93.2                                & \textbf{94.9}                        & \textbf{94.9}                      & 94.5                                                                                   & 94.5                            \\
                                & \textbf{F1}          & ~~96.5                                & 84.8                                 & 95.9                               & \textbf{96.5}                                                                          & \textbf{96.5}                   \\
                                & \textbf{Acc.}    & ~~93.2                                & 74.3                                 & 92.4                               & \textbf{93.3}                                                                          & \textbf{93.3}                   \\ \midrule
\multirow{4}{*}{\rotatebox[origin=c]{90}{\textbf{Chinese}}}    & \textbf{R.}      & 100.0                               & 82.3                                 & 98.7                               & 98.8                                                                                   & \textbf{99.6}                   \\
                                & \textbf{P.}   & ~~87.9                                & 89.9                                 & \textbf{90.1}                      & \textbf{90.1}                                                                          & 89.7                            \\
                                & \textbf{F1}          & ~~93.5                                & 85.9                                 & 94.2                               & 94.2                                                                                   & \textbf{94.4}                   \\
                                & \textbf{Acc.}    & ~~87.9                                & 76.3                                 & 89.3                               & 89.4                                                                                   & \textbf{89.5}                   \\ \midrule
\multirow{4}{*}{\rotatebox[origin=c]{90}{\textbf{Total}}} & \textbf{R.}      & 100.0                               & 79.1                                 & 97.7                               & 98.7                                                                                   & \textbf{98.9}                   \\
                                & \textbf{P.}   & ~~90.8                                & 92.6                                 & \textbf{92.8}                      & 92.6                                                                                   & 92.4                            \\
                                & \textbf{F1}          & ~~95.2                                & 85.3                                 & 95.2                               & \textbf{95.5}                                                                          & \textbf{95.5}                   \\
                                & \textbf{Acc.}    & ~~90.8                                & 75.2                                 & 91.0                               & \textbf{91.6}                                                                          & \textbf{91.6}         \\ \bottomrule \bottomrule 
\end{tabular}}
\caption{Performance for the Triplet Critic. \textbf{All True} shows baseline when all triplets are marked as True. \textbf{XNLI} uses the entailment prediction of a model trained solely on XNLI. \textbf{Ours} is trained solely on our annotated data without ar/zh. Last two columns show the multi-task approach with (\textbf{+XNLI}) and without ar/zh (\textbf{+XNLI$^{-}$}) data from XNLI.}
\label{tab:results_critic}
\end{table}

\subsection{Triplet Critic}
\label{sec:critic}
Our manual annotation procedure (Section \ref{sec:annotation}) filtered a portion of the silver data in order to have a higher-quality subset on which to train and evaluate our models. However, by removing the negative triplets we disregard valuable information that can be used to improve the quality of the remaining annotations, i.e. all those not validated by humans. Inspired by \citet{https://doi.org/10.48550/arxiv.2110.07178}, who trained \textit{critics} based on human annotations on commonsense triplets, we use our annotated triplets, both true and false positives with their contexts, to train a Triplet Critic. Specifically, given a textual context $c$ and an annotated triplet $t$ that may appear in $c$, we train a cross-encoder $T(c, t)$ to predict whether $c$, the premise, entails $t$, the hypothesis. 
This setup was inspired by NLI and results in training examples such as:

\begin{table}[h!]
\resizebox{\columnwidth}{!}{
\begin{tabular}{lcc}
\multicolumn{1}{c}{\textbf{Premise}}                                                                                                                                    & \multicolumn{1}{c}{\textbf{Hypothesis}}                                                          & \multicolumn{1}{c}{\textbf{Label}} \\ 
\multirow{3}{*}{\begin{tabular}[c]{@{}l@{}}Telefe (acronym for \\ Televisión Federal) \\ is a television station \\ located in Buenos\\ Aires, Argentina.\end{tabular}} & \begin{tabular}[c]{@{}c@{}}Telefe instance of \\ television station\end{tabular} & True           \\ \cline{2-3} 
                                                                                                                                                                        & \begin{tabular}[c]{@{}c@{}}Buenos Aires \\ country Argentina\end{tabular}        & True           \\ \cline{2-3} 
                                                                                                                                                                        & \begin{tabular}[c]{@{}c@{}}Argentina capital \\ Buenos Aires\end{tabular}        & False          \\ 
\end{tabular}}
\end{table}
Once $T$ is trained, we can use it on our silver data to filter out other false positives, i.e., triplets that, albeit present in Wikidata for two entities within the context, are not entailed by that context. 

We test our approach by training $T$ in English, French, Italian, Spanish and German, namely European languages with shared families (Romance and Germanic), and testing on Arabic and Chinese. This setup will test the zero-shot multilingual capabilities on unseen languages in order to determine whether the Critic can be applied to any language.
We base our Triplet Critic on DeBERTaV3 \cite{https://doi.org/10.48550/arxiv.2111.09543} with a classification head on top of the [CLS] token that produces a binary prediction, trained using a Cross-Entropy loss criterion. Furthermore, since the task is similar to and inspired by NLI, we explore a multi-task approach using the XNLI dataset \cite{conneau-etal-2018-xnli}, aiming at improving cross-lingual performance. To this end, we add an additional linear layer at the end of the model for NLI that projects the output layer to the three possible predictions (neutral, contradiction, entailment), again using a Cross-Entropy loss. 

Table \ref{tab:results_critic} shows the Triplet Critic results when trained under different setups. We see how the use of our data  dramatically improves upon the XNLI baseline, especially in terms of accuracy. While we are primarily interested in precision so as to guarantee that triplets are valid, a low accuracy would lead to missing annotations, which we also want to avoid. Additionally, when our Triplet Critic is trained simultaneously on our data and XNLI, even if no Arabic or Chinese data is used (i.e. +XNLI$^{-}$), performance further improves: the system achieves an average 92.6\% precision, on par with using only our data, but sees a point increase in recall, which is remarkable, taking into account the high class inbalance. In contrast, when XNLI data from those languages is added (i.e. +XNLI), we observe a small trade-off between precision and recall.

Overall, these zero-shot results certainly legitimize the use of our Triplet Critic to refine our silver data for unseen languages, and suggest even more promising benefits for seen languages. Furthermore, the Triplet Critic serves as a feedback on the consistency of human annotations since the models have successfully learned from them.

\subsection{Entity Typing}\label{sec:entity-typing}
In RE datasets, the entity types are commonly included in the triplets \cite{10.1007/978-3-642-15939-8_10} and, therefore, are taken into account under the strict evaluation, where a triplet is only considered correctly extracted when entity boundaries, entity types, and relation type are all predicted, versus the boundaries evaluation, where only entity boundaries and relation type are taken into account \cite{taille-etal-2020-lets}.
This Section describes the procedure through which we automatically label entities with their types. 

We start by mapping entities in Wikipedia to BabelNet \cite{NAVIGLI2012217,naviglietal-2021} synsets by exploiting the one-to-one linkage between them. Then, we annotate synsets by applying the knowledge-based semantic classifier introduced by \citet{tedeschi-etal-2021-wikineural-combined}, which exploits the relational information in BabelNet such as hypernymy and hyponymy relations. This procedure yields $\sim$7.5M entities labeled with an entity type. However, since the annotations are automatically derived and prone to errors, we devise a new strategy to improve their quality. Specifically, we design a Transformer-based classifier that takes a synset and returns its NER category. More formally, let us define the functions $L(s)$ and $D(s)$ that output the main lemma and the textual description of a synset $s$, respectively. Then, given a synset $s$ and the above-defined functions, we provide the string $I(s, D, L)$ = [CLS] $L(s)$ [SEP] $D(s)$ [SEP] as input to the classifier that predicts a label e $\in$ E\footnote{E = \{location, person, number, time, organization, date, event, celestial body, media, disease, concept, miscellaneous and unknown\}}. The tagset E is obtained by refining the categorization of named entities introduced by \citet{tedeschi-etal-2021-named-entity} based on the ability of automated systems to distinguish NER categories and on the frequency of these categories in Wikipedia articles \cite{tedeschi-navigli-2022-multinerd}. 
To train the classifier, we construct a dataset by selecting a high-quality subset from the 7.5M automatically-produced annotations by taking only synsets with a maximum distance equal to 1 from one of the 40k synsets in WordNet \cite{Milleretal:90}, this latter being a manually-curated subset of BabelNet. By doing this, we obtain a set consisting of 1.2M high-quality annotations that we split into 80\% for training and 20\% for validation, and convert these to the above-specified $I(s, D, L)$ format.

Finally, we use the trained classifier to confirm or replace the previous 7.5M annotations, resulting in 6.2M (82.4\%) confirmations and 1.3M (17.6\%) changes, and employ it to label new Wikidata instances as well, thus obtaining a final mapping consisting of $\sim$13M Wikidata entries annotated with their entity types. By manually inspecting a sample of 100 changes, we observed that our NER classifier was right 68\% of the time, wrong 17\%, while the remaining 15\% of the time both annotations were wrong, providing an improvement of 51\% over 1.3M changes. 
We highlight that 68\% is not the accuracy of our classifier as it is computed on 100 items where there is a disagreement between the original annotations produced by  WikiNEuRal \cite{tedeschi-etal-2021-wikineural-combined} --the current state of the art in entity typing-- and the annotations produced by our model. Indeed, an accuracy of 68\% on this subset means that our classifier corrected most of the instances that were previously mistaken by WikiNEuRal. For completeness, we report that when the two systems agree, i.e. 82\% of the time, they are correct in 98\% of these cases, as measured on another subset of 100 instances.

\begin{table}[t]
\centering
\resizebox{\columnwidth}{!}{
\begin{tabular}{rrrrrrrrr}
\toprule
\multicolumn{1}{l}{} & \multicolumn{4}{c}{Human Annotated}               &                                                                                                                                                                                                                                                                                                                                                                                                                            \multicolumn{4}{c}{Distant supervision}                                                                      \\ \cmidrule(lr){2-5}   \cmidrule(lr){6-9} 
\textbf{}            & \multicolumn{1}{c}{\rotatebox[origin=c]{90}{ACE05}} & \multicolumn{1}{c}{\rotatebox[origin=c]{90}{CONLL04}} & \multicolumn{1}{c}{\rotatebox[origin=c]{90}{DocRED}} &  \multicolumn{1}{c}{\rotatebox[origin=c]{90}{RED\textsuperscript{FM}}}  & \multicolumn{1}{c}{\rotatebox[origin=c]{90}{DocRED}} & \multicolumn{1}{c}{\rotatebox[origin=c]{90}{NYT}} & \multicolumn{1}{c}{\rotatebox[origin=c]{90}{SMILER}} & \multicolumn{1}{c}{\rotatebox[origin=c]{90}{SRED\textsuperscript{FM}}} \\ \midrule
\textbf{Docs.}       & 1.6K                                                                             & 1.4K                                                                                                                                & 5K                                                                                & 15.4K                                                                                                                                    & 100K                         & 66.2K                      & 1.1M                        & 12.3M                      \\
\textbf{Sents.}       & 30.9K                                                                             & 1.4K                                                                                                                                & -                                                                                & 43.7K                                                                                                                                    & -                         & 66.2K                      & 1.1M                        & 46.6M                      \\
\textbf{Rels.}       & 6                                                                             & 5                                                                                                                                & 96                                                                                & 32                                                                                                                                    & 96                         & 24                      & 36                        & 400                      \\
\textbf{Ents.}       & 5                                                                             & 4                                                                                  & 6                                                                                 & 13                                                                                                                            & 6                          & 3                       & -                          & 13                        \\ \hline
\textbf{AR}          & 4.7K                                                                          & -                                                                                                       & -                                                                                 & 1.8K                                                                                                                    & -                          & -                       & 9K                         & 3.3M                     \\
\textbf{CA}          & -                                                                             & -                                                                            & -                                                                     & -                                                                     & -                          & -                       & -                          & 1.7M                     \\
\textbf{DE}          & -                                                                             & -                                                 & -                                                                                 & 7.5K                                                                                                                              & -                          & -                       & 53K                        & 4.8M                     \\
\textbf{EL}          & -                                                                             & -                                                                                                                           & -                                                                     & -                                                                     & -                          & -                       & -                          & 325K                     \\
\textbf{EN}          & 8.7K                                                                                       & 6.8K                                                                  & 58.6K                                                                             & 10.9K                                                                                                                         & 1M                         & 111K                    & 748K                       & 12.4M                    \\
\textbf{ES}          & -                                                                             & -                                             & -                                                                                 & 6.5K                                                                                                                          & -                          & -                       & 12K                        & 4.2M                     \\
\textbf{FA}          & -                                                                             & -                                                                                  & -                                                        & -                                                                     & -                          & -                       & 3K                         & -                        \\
\textbf{FR}          & -                                                                             & -                                                                                  & -                                                      & 7.4K                                                                                                                      & -                          & -                       & 62K                        & 4.2M                     \\
\textbf{HI}          & -                                                                             & -                                                                                  & -                                                       & -                                                                                                                                & -                          & -                       & -                          & 301K                     \\
\textbf{IT}          & -                                                                             & -                                                                                  & -                                                        & 6.8K                                                                                                                               & -                          & -                       & 76K                        & 2M                       \\
\textbf{JA}          & -                                                                             & -                                                                                  & -                                                       & -                                                                                                                           & -                          & -                       & -                          & 3.3M                     \\
\textbf{KO}          & -                                                                             & -                                                                                  & -                                                       & -                                                                                                                                      & -                          & -                       & 20K                        & 1M                       \\
\textbf{NL}          & -                                                                             & -                                                                                  & -                                                       & -                                                                                                                                    & -                          & -                       & 40K                        & 3M                       \\
\textbf{PL}          & -                                                                             & -                                                                                  & -                                                       & -                                                                                                                                   & -                          & -                       & 17K                        & 3.7M                     \\
\textbf{PT}          & -                                                                             & -                                                                                  & -                                                       & -                                                                                                                                  & -                          & -                       & 45K                        & 2.7M                     \\
\textbf{RU}          & -                                                                             & -                                                                                  & -                                                       & -                                                                                                                              & -                          & -                       & 7K                         & 1.6M                     \\
\textbf{SV}          & -                                                                             & -                                                                               & -                                                                     & -                                                                     & -                          & -                       & 5K                         & 7.2M                     \\
\textbf{UK}          & -                                                                             & -                                                                      & -                                                                     & -                                                                     & -                          & -                       & 1K                         & -                        \\
\textbf{VI}          & -                                                                             & -                                                                                & -                                                                     & -                                                                     & -                          & -                       & -                          & 1.4M                     \\
\textbf{ZH}          & 9.3K                                                                          & -                                                      & -                                                                                 & 1.4K                                                                                                                             & -                          & -                       & -                          & 3M                       \\ \hline
\end{tabular}
}
\caption{Number of relation types (Rels.), entity types (Ents.) and annotated triplets in RE resources.}
\label{tab:numbers}
\end{table}
\subsection{SRED\textsuperscript{FM}}
\label{sec:comparison}

\begin{figure*}[t!]
    \centering
    \includegraphics[width=\textwidth]{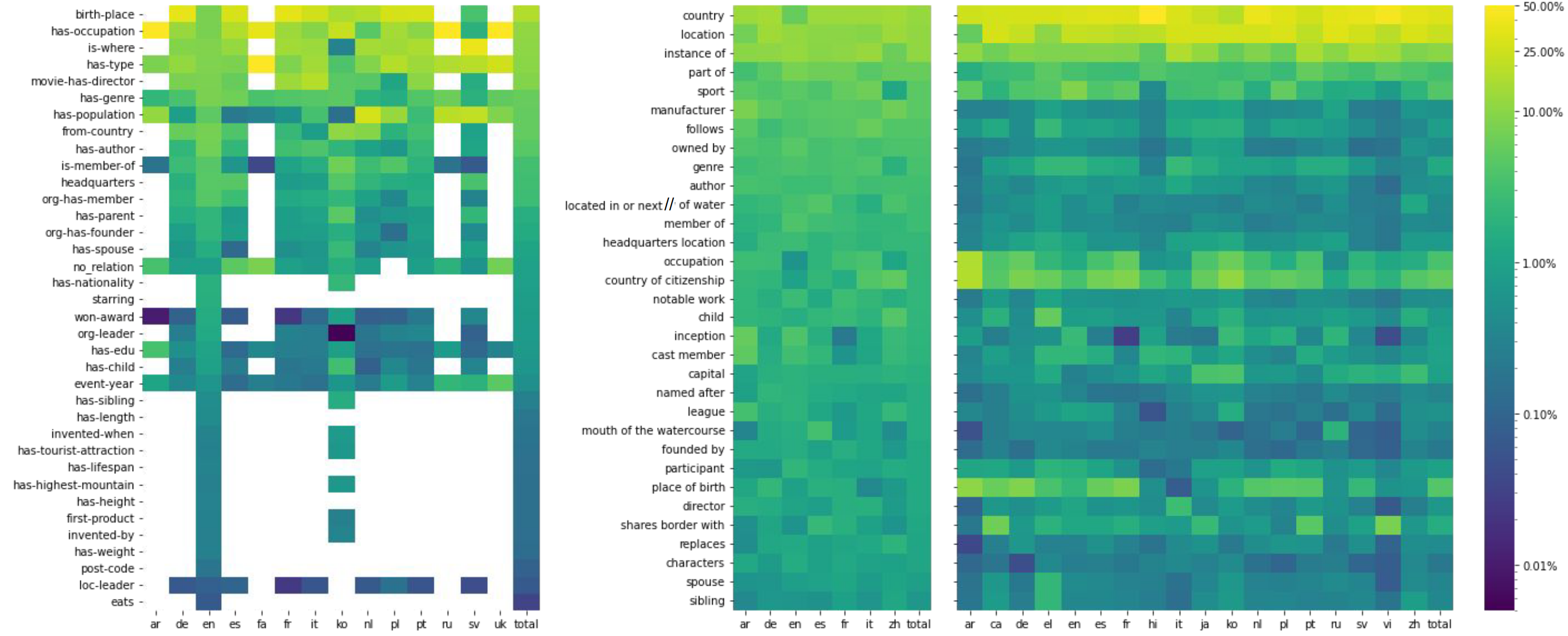}    
    \caption{Comparison of relation type distribution by percentage between SMiLER (left),  RED\textsuperscript{FM} and SRED\textsuperscript{FM} (right). Best seen in color.} 
    \label{fig:stats}
\end{figure*}
The current datasets for Relation Extraction often lack complete coverage of relations. The SMiLER dataset \cite{seganti-etal-2021-multilingual} only annotates one triplet per example, resulting in a limited understanding of the relationships therein. For instance, in the example "\textit{Fredrik Hermansson (born 18 July 1976) is a Swedish musician. He was a keyboardist and backing vocalist in the Swedish progressive rock band Pain of Salvation.}", the triplet (Fredrik Hermansson, has-genre, progressive rock) is annotated, but other triplets such as (Fredrik Hermansson, has-occupation, musician) and (Fredrik Hermansson, has-nationality, Swedish) are also valid.

Another common issue with RE datasets is the high class imbalance, particularly for distantly annotated datasets. This is often due to the skewed distributions that are intrinsic in knowledge bases such as Wikidata, which are often used to construct RE resources. This leads to low coverage for lower frequency classes, as seen in Figure \ref{fig:stats} for certain languages in the SMiLER dataset. In DocRED \cite{yao-etal-2019-docred}, another distantly annotated dataset, location-based relations constitute over 50\% of instances.

Fully human-annotated datasets that overcome these limitations are scarce and often not widely accessible. Additionally, they often cover a narrow set of languages and relations (Table \ref{tab:numbers}). To address these issues, we introduce Silver RED\textsuperscript{FM}. SRED\textsuperscript{FM} is a large, multilingual RE dataset that contains more than 45M triplets and covers 400 relation types and 18 languages. It is created using the data extraction procedure described in Section \ref{sec:data-extraction} and the Triplet Critic introduced in Section \ref{sec:critic}. SRED\textsuperscript{FM} overcomes some of the previous shortcomings of current datasets by providing a higher coverage of annotation and more evenly distributed classes. 

Using the same example sentence, in SRED\textsuperscript{FM}, the following triplets are annotated: (Fredrik Hermansson, country of citizenship, Swedish), (Fredrik Hermansson, occupation, musician), (Fredrik Hermansson, date of birth, 18 July 1976), and (Fredrik Hermansson, member of, Pain of Salvation). 
Regarding class balance, in the English portion of SRED\textsuperscript{FM} location-based relations make up less than 37\%. Hence our datasets have more evenly distributed classes. Figure \ref{fig:stats} (right) shows the distribution for the top 32 of the 400 relation types in SRED\textsuperscript{FM}.

Additionally, we provide a pipeline that enables the automatic creation of an RE dataset in any language. So, even though we release the SRED\textsuperscript{FM} dataset as described in this paper (i.e. covering 18 different languages), we encourage the expansion to other languages by using our pipeline available \href{https://github.com/Babelscape/crocodile}{here}.

In summary, SRED\textsuperscript{FM} is a large, multilingual dataset that addresses the shortcomings of current datasets by providing a higher coverage of annotation and more evenly distributed classes. RED\textsuperscript{FM}, instead, is the result of the manual annotation (Section \ref{sec:data-extraction}) to which we add entity types. We split them into training, validation and test, with no overlapping Wikipedia pages across splits. Details can be found in Table \ref{tab:REDFM} in Appendix \ref{sec:appendix_data}.

\section{mREBEL}
\begin{table*}[t]
\resizebox{2\columnwidth}{!}{
\begin{tabular}{lllll}
\toprule
                             &  & \textbf{Input text}                                                                                                                                                                                                                                                                                             & \textbf{Triplets}                                                                                                                                                                                                                                                                                   & \textbf{Linearized Triplets}                                                                                                                                                                                                                                                                                                                                                                                                                                                                                                                                                                                                                       \\ \midrule
\multirow{1}{*}[5ex]{\rotatebox[origin=c]{90}{\textbf{Classification}}}
                             & nl             & \begin{tabular}[c]{@{}l@{}}nl\_XX \# Mumbai Mirror \# is een Engelstalige tabloid, \\ die verschijnt in de Indiase stad Mumbai. \\ Het is hier met een oplage van zo'n 700.000 \\ exemplaren de belangrijkste krant. Het dagblad \\ verscheen voor het eerst op @ 30 mei 2005 @,\end{tabular}                   & \begin{tabular}[c]{@{}l@{}}(Mumbai Mirror, \\ inception, \\ 30 mei 2005)\end{tabular}                                                                                                                                                                                                               & \begin{tabular}[c]{@{}l@{}}tp\_XX\textless{}relation\textgreater Mumbai Mirror \\ \textless{}media\textgreater 30 mei 2005 \textless{}date\textgreater inception\end{tabular}                                                                                                                                      \\ \midrule
\multirow{1}{*}[4ex]{\rotatebox[origin=c]{90}{\textbf{Extraction}}} & ca           & \begin{tabular}[c]{@{}l@{}}ca\_XX Can Verboom és una masia amb elements \\ gòtics i barrocs de Premià de Dalt ( Maresme ) \\ protegida com a bé cultural d'interès local.\end{tabular}                                                                                                                          & \begin{tabular}[c]{@{}l@{}}(Can Verboom, \\ located in the \\administrative territorial entity, \\ Premià de Dalt)\\ (Can Verboom, \\heritage designation, \\ bé cultural d'interès local)\end{tabular}                                                                                                 & \begin{tabular}[c]{@{}l@{}}tp\_XX\textless{}triplet\textgreater Can Verboom \textless{}loc\textgreater \\ Premià de Dalt \textless{}loc\textgreater located in the \\administrative territorial entity  \textless{}loc\textgreater \\ bé cultural d'interès local \textless{}loc\textgreater \\ heritage designation\end{tabular}                                                                                                                                                                                                                                                                                                                     \\
 \bottomrule
\end{tabular}}
\caption{Examples from SRED\textsuperscript{FM} with their triplet linearization used to train mREBEL.}
\label{tab:mREBEL_examples}
\end{table*}
\begin{table*}[t]
\centering
\resizebox{\textwidth}{!}{%
\begin{tabular}{lrrrrrrrrrrrrrrr}\toprule
\textbf{}      & \textbf{ar}   & \textbf{de}   & \textbf{en}   & \textbf{es}   & \textbf{fa}   & \textbf{fr}   & \textbf{it}   & \textbf{ko}   & \textbf{nl}   & \textbf{pl}   & \textbf{pt}   & \textbf{ru}   & \textbf{sv}   & \textbf{uk}   & \textbf{Avg.} \\ \midrule
Support         & 190           & 1049          & 731           & 225           & 54            & 1241          & 1501          & 228           & 791           & 343           & 880           & 131           & 92            & 20            &                  \\ \midrule
HERBERTa         & 49.0          & 63.0          & 77.0          & 46.0          & 72.0          & 58.0          & 69.0          & 51.0          & 61.0          & 50.0          & 62.0          & 30.0          & 59.0          & 25.0          & 54.7             \\
mREBEL$^{T}_{400}$   & \textbf{75.3} & 63.1          & \textbf{77.7} & \textbf{57.8} & 69.2          & 64.3 & \textbf{83.5} & \textbf{62.5} & 72.8          & 76.1          & 69.8          & 71.0          & 76.1          & \textbf{70.0} & 70.5       \\ 
mREBEL$_{400}$ & 73.7          & \textbf{64.2}          & 77.7          & 54.7          & \textbf{74.8}          & \textbf{66.1}          & 82.9          & 61.7          & \textbf{73.6} & \textbf{76.4} & \textbf{70.1} & \textbf{75.6} & \textbf{78.3} & 65.0          & \textbf{70.8}             \\
\midrule
mT5$_{BASE}$* & 95.1          & 95.4          & 96.1          & 81.1          & 73.1           & 97.2          & 98.3 & 83.2          & 96.9          & 95.6          & 96.9          & 87.6           & 63.0          & 71.8           & 88.0          \\
mT5$_{BASE}$(en)  & 94.0            & 94.9          & -             & 91.7          & 91.1           & 96.0            & 97.5          & 78.2          & 97.5          & 93.3          & 95.2          & 93.8           & 97.8          & 94.7           & 93.5          \\
mREBEL$^{T}_{B400}$                 & \textbf{99.5} & 96.8          & 96.6          & 95.1          & \textbf{100.0} & 97.7          & 98.7          & \textbf{94.7} & 98.3          & 97.4          & 97.8          & \textbf{100.0} & 96.7          & \textbf{100.0} & 97.8          \\
mREBEL$^{T}_{400}$                  & \textbf{99.5} & 97.4          & 97.5          & 94.9          & 97.0           & \textbf{97.8}          & 98.8          & 93.1          & \textbf{98.7}          & 98.4          & 98.3          & \textbf{100.0} & 97.8          & \textbf{100.0} & 97.8          \\
mREBEL$^{T}_{400}$*           & \textbf{99.5} & 96.9          & 97.5          & 93.5          & 99.0           & \textbf{97.8}          & \textbf{98.9} & 94.1          & 98.0          & 98.3          & 97.7          & 98.8           & \textbf{98.4} & 97.4           & 97.6          \\
mREBEL$_{400}$                      & \textbf{99.5} & \textbf{97.5}          & \textbf{97.7} & \textbf{95.3} & \textbf{100.0} & 97.6          & \textbf{98.9} & \textbf{94.7} & 98.6          & \textbf{98.7} & \textbf{98.4} & 99.2           & 97.8          & \textbf{100.0} & \textbf{98.1}
\\
\bottomrule     
\end{tabular}%
}
\caption{Results on SMiLER. Micro-F1 scores per language. Top half shows RE, bottom half RC. Selected top performing multilingual HERBERTa per language from \cite{seganti-etal-2021-multilingual} and mT5$_{BASE}$ from \citet{chen-etal-2022-multilingual}. \citet{chen-etal-2022-multilingual}(en) was trained on English data. * indicates separate training per language.}
\label{tab:SMiLER-results}
\end{table*}
In this section, we present our system, mREBEL (Multilingual Relation Extraction By End-to-end Language generation), which is a multilingual relation extraction model pre-trained on SRED\textsuperscript{FM}. It is a multilingual extension of the REBEL model introduced in \citet{huguet-cabot-navigli-2021-rebel-relation}, which uses a seq2seq architecture to convert relations into text sequences that can be decoded by the model. 
We convert triplets into text sequences and pre-train our model using mBART-50 \cite{tang-etal-2021-multilingual}. To support multiple languages, we prepend the input text with a special language token (i.e. \texttt{en\_XX}). Additionally, we include relation classification (RC) in the pre-training phase of mREBEL. Specifically, for 5\% of the training data, we select a random triplet, mark the subject and object entities in the input text, and use a special token \texttt{<relation>} to indicate to the model that only one triplet needs to be decoded. Finally, to promote cross-lingual transfer, we use the English names for the relation types when decoding the triplets. Table \ref{tab:mREBEL_examples} shows how instances from SRED\textsuperscript{FM} are used to train mREBEL.\\
We train three versions of mREBEL:
\begin{enumerate}
    \item mREBEL$^{T}_{400}$, trained on 400 relation types, including entity types;
    \item mREBEL$^{T}_{32}$, fine-tuned on top of the previous one but including only the 32 relation types from RED\textsuperscript{FM};\item mREBEL$^T_{B400}$, trained on top of M2M100 \cite{fan-etal-2020-beyond}.
\end{enumerate}
For (1) and (2) we also train their untyped versions, mREBEL$_{400}$ and mREBEL$_{32}$.

\section{Experimental Setup}
We evaluate mREBEL and its variants on our own datasets, i.e. RED\textsuperscript{FM} and SRED\textsuperscript{FM}, and on SMiLER \cite{seganti-etal-2021-multilingual}. 
Unless stated otherwise, we train on the training sets of all languages simultaneously and apply early stopping based on the Micro-F1 obtained on the overall validation set. We use the Adafactor optimizer and the same Cross-Entropy loss with teacher forcing from \citet{huguet-cabot-navigli-2021-rebel-relation}. The full list of hyperparameters is detailed in Appendix \ref{app:repro}.

\begin{table*}[t]
\centering
\begin{tabular}{lc|ccccccc|ccc}\toprule
\multicolumn{1}{l}{\textbf{Model}}        & \textbf{Fine-tuning} & \textbf{ar} & \textbf{de} & \textbf{en} & \textbf{es} & \textbf{fr} & \textbf{it} & \textbf{zh} & \textbf{P} & \textbf{R} & \textbf{F1} \\ \midrule
mREBEL$^{T\dagger}_{32}$ &  \xmark                        & 43.4        & 53.9        & 50.0        & 46.9        & 48.3        & 56.4        & 38.5        & 43.1       & 56.1       & 48.7        \\
mREBEL$^{T}_{32}$  & \xmark                          & \textbf{45.5}        & 57.8        & 53.3        & \textbf{51.4}        & \textbf{52.5}        & \textbf{57.8}        & \textbf{41.8}        & 47.6       & \textbf{57.3}       & \textbf{52.0}        \\
mBART  & \cmark                         & 16.1        & 39.6        & 32.6        & 27.3        & 28.5        & 30.0        & ~~0.0        & 26.8       & 29.0       & 27.9        \\
mREBEL$^{T}_{B400}$ &  \cmark                         & 39.9        & 50.3        & 49.0        & 41.1        & 41.7        & 50.6        & 38.0        & 45.0       & 45.1       & 45.1        \\
mREBEL$_{400}$ &  \cmark                         & 33.2        & 50.0        & 42.2        & 38.4        & 40.3        & 49.2        & 30.7        & 40.9       & 41.7       & 41.3        \\
mREBEL$^{T}_{400}$ & \cmark       & 39.3        & 52.8        & 49.5        & 45.9        & 46.8        & 54.7        & 35.2        & 47.5       & 47.0       & 47.2        \\
mREBEL$^{T\dagger}_{32}$ & \cmark & 43.7        & 55.4        & \textbf{54.0}        & 46.9        & 50.5        & 57.1        & 38.8        & 47.9       & 53.0       & 50.3        \\
mREBEL$^{T}_{32}$& \cmark        & 43.8        & \textbf{58.3}        & 53.7        & 50.1        & 51.8        & \textbf{57.8}        & 41.3        & \textbf{48.9}       & 54.7       & 51.6       \\ \bottomrule
\end{tabular}
\caption{Results on RED\textsuperscript{FM} test set. Micro-F1 scores per language. $\dagger$ indicates the Critic was not used to filter pre-training data. Fine-tuning indicates fine-tuning on RED\textsuperscript{FM} training set.}
\label{tab:redfm-results}
\end{table*}

\begin{table*}[t]
\centering
\resizebox{\textwidth}{!}{%
\begin{tabular}{lccccccccccccccccccc}\toprule
\textbf{}          & \textbf{ar} & \textbf{ca} & \textbf{de} & \textbf{el} & \textbf{en} & \textbf{es} & \textbf{fr} & \textbf{hi} & \textbf{it} & \textbf{ja} & \textbf{ko} & \textbf{nl} & \textbf{pl} & \textbf{pt} & \textbf{ru} & \textbf{sv} & \textbf{vi} & \textbf{zh} & \textbf{all} \\ \midrule
\textbf{Sents.} (K) & ~~4.5        & ~~2.7        & ~~5.7        & ~~0.9         & ~~6.9        & ~~8.8        & ~~4.0        & ~~0.4         & ~~2.2        & ~~3.4        & ~~0.8         & ~~2.8        & ~~6.5        & ~~4.5        & ~~2.3        & ~~7.6        & ~~2.0        & ~~2.7        & 68.6        \\
\textbf{Trip.} (K) & 27.9       & 15.5       & 32.1       & ~~4.9        & 40.0      & 54.9       & 26.1       & ~~2.8        & 12.8       & 29.4       & ~~4.4        & 16.4       & 36.6       & 27.5       & 12.9      & 44.9       & 10.5       & 27.4       & 427.0~~       \\
\textbf{Precision} & 61.1     & 57.0     & 58.5     & 46.1     & 62.3     & 60.6     & 60.0     & 50.2     & 57.7     & 48.6     & 48.5     & 57.5     & 64.6     & 60.0     & 53.4     & 63.2     & 59.0     & 41.0     & 58.5      \\
\textbf{Recall}    & 48.7     & 45.9     & 44.5     & 34.1     & 47.2     & 51.2     & 45.9     & 29.6     & 44.9     & 44.2     & 37.7     & 49.0     & 55.4     & 48.9     & 38.2     & 55.7  & 45.3     & 36.6     & 47.9      \\
\textbf{Micro-F1}  & 54.2     & 50.8     & 50.6     & 39.2     & 53.7     & 55.5     & 52.0     & 37.2     & 50.5     & 46.3     & 42.5     & 52.9     & 59.6     & 53.9     & 44.6     & 59.2     & 51.2     & 38.7     & 52.7     \\ 
\textbf{Macro-F1}    & 24.0     & 24.8     & 29.3     & 12.5 & 36.3     & 30.5     & 29.1     & ~~8.3     & 27.9     & 25.1     & 17.3     & 27.6    & 30.5     & 29.5     & 23.5     & 30.1  & 19.5     & 20.3     & 24.8      \\
\bottomrule
\end{tabular}%
}
\caption{Results for SRED\textsuperscript{FM} test set with mREBEL$_{400}^T$ on 400 relation types.}
\label{tab:sred_results}
\end{table*}
\paragraph{Multilingual Relation Extraction} 

We report the Micro-F1 score per language for both SMiLER and RED\textsuperscript{FM} test sets. When evaluating on SMiLER, we use mREBEL$_{400}$ variants as starting checkpoints and fine-tune them on SMiLER training sets. For RED\textsuperscript{FM}, instead, we include the mREBEL$_{32}$ model in our experiments as it was trained on the same set of relations. The inclusion of this model lets us analyze the impact of further fine-tuning on RED\textsuperscript{FM} gold data against the quality of our silver annotation process.
As an extrinsic evaluation of our Triplet Critic model from Section \ref{sec:critic}, we train a version of mREBEL$^{T}_{32}$ without filtering triplets. 

\paragraph{Multilingual Relation Classification} Even though \citet{seganti-etal-2021-multilingual} introduced SMiLER as a RE dataset, each sentence contains just one annotated triplet and includes the ``no relation'' class as part of its annotation scheme. Therefore, it is better approached as an RC task and it is more akin to a dataset like TACRED. For  instance, \citet{chen-etal-2022-multilingual} use it as an RC dataset with an array of prompt-based approaches, and we compare our approach with theirs for RC. 

\section{Results}

\begin{table*}[t]
\centering
\resizebox{\textwidth}{!}{
\begin{tabular}{lll}\toprule
\textbf{Error Type} & \textbf{Count} & \textbf{Example}  \\ \midrule\midrule
Entity type         & 262 (7.2\%)     & 
\begin{dependency}[label style={font=\bfseries\large,inner sep=0.75ex}, baseline={(0,0.1)}, baseline={(0,0.1)}]
   \begin{deptext}[column sep=.1cm, row sep=.2ex]
      The \& Peugeot \& 408 \& is a small family car produced by \& Peugeot \& . \\
       \& CONCEPT  \& MISC\&  \& ORG \& \\
    \end{deptext}
   \depedge[edge start x offset=20pt,edge unit distance=0.8ex, edge style={ultra thick}]{2}{5}{manufacturer}
     \wordgroup[group style={draw=gray, inner sep=.05ex}]{1}{2}{3}{a3}
   \wordgroup[group style={fill=orange!40, draw=brown, inner sep=.05ex}]{2}{2}{2}{a0}
      \wordgroup[group style={fill=green!40, draw=green, inner sep=.05ex}]{2}{3}{3}{a1}
      \wordgroup[group style={fill=green!40, draw=green, inner sep=.05ex}]{2}{5}{5}{a2}
      \wordgroup[group style={draw=gray, inner sep=.05ex}]{1}{5}{5}{a4}
\end{dependency}
\\ \midrule
Span underlap       & 157 (4.3\%)     & 
\begin{dependency}[label style={font=\bfseries\large,inner sep=0.75ex}, baseline={(0,0.15)}]
   \begin{deptext}[column sep=.1cm, row sep=.5ex]
      La \& Changhe Aircraft Industries \& Corporation \& è un'azienda cinese, specializzata nella produzione di \& elicotteri \& .\\
       \& ORG \&  \&  \& CONCEPT \& \\
 \end{deptext}
   \depedge[edge start x offset=20pt, edge unit distance=0.6ex, edge vertical padding=0.6ex, edge style={ultra thick}]{2}{5}{manufacturer}
   \wordgroup[group style={fill=green!40, draw=green, inner sep=.5ex}]{1}{2}{3}{a1}
   \wordgroup[group style={fill=orange!40, draw=brown, inner sep=.01ex}]{1}{2}{2}{a0}

     \wordgroup[group style={draw=gray, inner sep=.5ex}]{1}{5}{5}{a3}
\end{dependency}
\\ \midrule
Span overlap        & 138 (3.8\%)     & \begin{dependency}[label style={font=\bfseries\large,inner sep=0.75ex}, baseline={(0,0.15)}]
   \begin{deptext}[column sep=.05cm, row sep=.5ex]
      La caja de Pandora \&, \& película \&[-.5cm] muda \&[-.3cm] alemana, dirigida por G. W. Pabst. \\
       MEDIA \&  \&  \&[-0.4cm] CONCEPT  \& \\
 \end{deptext}
   \depedge[edge end x offset=20pt, edge unit distance=1ex, edge vertical padding=0.6ex, edge style={ultra thick}]{1}{3}{genre}
    \wordgroup[group style={fill=orange!40, draw=brown, inner sep=.5ex}]{1}{3}{4}{a0}
   \wordgroup[group style={fill=green!40, draw=green, inner sep=.01ex}]{1}{4}{4}{a1}
     \wordgroup[group style={draw=gray, inner sep=.5ex}]{1}{1}{1}{a3}
\end{dependency}
\\ \midrule
Subject             & 269 (7.4\%)     & 
\begin{dependency}[label style={font=\bfseries\large,inner sep=0.75ex}, baseline={(0,0.1)}]
   \begin{deptext}[column sep=.1cm, row sep=.2ex]
      L'œuvre contient notamment " \&Le Pré Béjine \& ", dont le titre a été repris par \& Sergueï Eisenstein \& pour un \& film inachevé \&.\\
        \& MEDIA \&  \& PERSON  \& \& MEDIA \& \\
 \end{deptext}
   \depedge[edge unit distance=1ex, edge style={ultra thick}]{2}{4}{director}
   \depedge[edge unit distance=1ex, edge style={ultra thick}]{6}{4}{director}
    \wordgroup[group style={fill=orange!40, draw=brown, inner sep=.01ex}]{1}{2}{2}{a1}
   \wordgroup[group style={fill=green!40, draw=green, inner sep=.01ex}]{1}{6}{6}{a0}
     \wordgroup[group style={draw=gray, inner sep=.01ex}]{1}{4}{4}{a3}
\end{dependency}
\\ \midrule
Object              & 117 (3.2\%)     &
\begin{dependency}[label style={font=\bfseries\large,inner sep=0.75ex}, baseline={(0,0.1)}]
   \begin{deptext}[column sep=.1cm, row sep=.2ex]
      OpenSolaris \& ist ein auf \& Unix \& basierendes \& Betriebssystem \& für die Plattformen PC, SPARC und andere. \\
       MISC \&  \& CONCEPT \&   \& CONCEPT \&  \& \\
 \end{deptext}
   \depedge[edge unit distance=1ex, edge style={ultra thick}]{1}{3}{instance of}
   \depedge[edge unit distance=1ex, edge style={ultra thick}]{1}{5}{instance of}
    \wordgroup[group style={fill=orange!40, draw=brown, inner sep=.01ex}]{1}{5}{5}{a1}
   \wordgroup[group style={fill=green!40, draw=green, inner sep=.01ex}]{1}{3}{3}{a0}
     \wordgroup[group style={draw=gray, inner sep=.01ex}]{1}{1}{1}{a3}
\end{dependency}          \\ \midrule
Relation            & 66   (1.8\%)    & 
\begin{dependency}[label style={font=\bfseries\large,inner sep=0.75ex}, baseline={(0,0.1)}]
   \begin{deptext}[column sep=.1cm, row sep=.2ex]
      The Red Hot Chili Peppers \& were formed in Los Angeles by Kiedis, Flea, guitarist \& Hillel Slovak \& and drummer Jack Irons. \\
       ORG \&  \& PER \&  \\
 \end{deptext}
    \depedge[edge unit distance=2.5ex, edge style={green!60!,ultra thick},
    label style={fill=green!40,text=black}]{1}{3}{member of}
   \depedge[edge unit distance=1ex, edge style={orange!60!,ultra thick},
    label style={fill=orange!40,text=black}]{1}{3}{part of}
     \wordgroup[group style={draw=gray, inner sep=.01ex}]{1}{1}{1}{a3}
     \wordgroup[group style={draw=gray, inner sep=.01ex}]{1}{3}{3}{a0}
\end{dependency}          \\ \midrule
Other               & 2606 (72.2\%)   & 
\begin{dependency}[label style={font=\bfseries\large,inner sep=0.75ex}, baseline={(0,0.1)}]
   \begin{deptext}[column sep=.1cm, row sep=.2ex]
      René L'Hermitte \& est un \& journaliste \& et un professeur des universités français, membre de  la Résistance et du Parti communiste \\
       PER \&  \& CONCEPT \& \\
 \end{deptext}
   \depedge[edge unit distance=1ex, edge style={orange!60!,ultra thick},
    label style={fill=orange!40,text=black}]{1}{3}{occupation}
     \wordgroup[group style={draw=gray, inner sep=.01ex}]{1}{1}{1}{a3}
     \wordgroup[group style={fill=orange!40, draw=brown, inner sep=.01ex}]{1}{3}{3}{a0}
\end{dependency}
\\
\bottomrule
\end{tabular}
}
\caption{Types of error encountered in RED\textsuperscript{FM}. Orange shows the mismatched prediction, and green is the annotated counterpart. Best seen in color.}
\label{tab:errors}
\end{table*}

\paragraph{Multilingual Relation Extraction} First, in Table \ref{tab:SMiLER-results} we show how our system performs compared to HERBERTa, the system proposed by \citet{seganti-etal-2021-multilingual} for SMiLER, using their best-performing setup for each language. 
We consider this dataset better suited for RC. However, as it originally reports on RE, we demonstrate how our system can perform better when pretrained on SRED\textsuperscript{FM}.
In particular, mREBEL$^{T}_{400}$ provides an improvement of about 15 Micro-F1 points compared to HERBERTa. Additionally, as SMiLER does not include entity types, we observe that mREBEL$_{400}$ performs marginally better than mREBEL$^{T}_{400}$. 

Table \ref{tab:redfm-results}, instead, shows the results on RED\textsuperscript{FM}, compared against an mBART baseline. Specifically, we analyze model performance when fine-tuning is, or is not, performed on the train set of RED\textsuperscript{FM}. While performances vary across languages, the best overall Micro-F1 (52.0) is obtained when training on SRED\textsuperscript{FM}, mREBEL$^{T}_{32}$, without further fine-tuning.  This confirms that our silver annotation procedure produces high-quality data, as there is no need for further tuning with RED\textsuperscript{FM}, which achieved 51.6. We also see how filtering by the Triplet Critic was crucial: when removed, performance dropped by more than 3 points. 

Training on 400 relation types does lead to lower results, since there is a mismatch between the two stages of training. However, mREBEL$^{T}_{400}$ showed decent performance on SRED\textsuperscript{FM} as shown in Table \ref{tab:sred_results}. This provides the first RE system to competitively extract up to 400 relation types in multiple languages. See Appendix \ref{app:extra_results} for more results. 

\paragraph{Multilingual Relation Classification} From the bottom half of Table \ref{tab:SMiLER-results}, we can observe how our mREBEL models 
consistently outperform competitive baselines, i.e. mT5$_{BASE}$* and mT5$_{BASE}$(en), by a large margin on all tested languages. 

\subsection{Error Analysis}
We performed an error analysis with mREBEL$^{T}_{32}$ to understand the sources of error when training on RED\textsuperscript{FM}. Our study revealed that 27.8\% of errors in the test set can be attributed to specific reasons.

First, there were discrepancies between predicted entity types and annotations (7.2\%). These errors may have arisen from the automatic nature of the typing annotation, errors by the system or ambiguity in some cases, such as fictional characters, which can be considered either PERSON or MEDIA. Additionally, a portion of errors (8.1\%) resulted from mismatches between the predicted and annotated spans for each entity, which may also be ambiguous (see the Span overlap example in Table \ref{tab:errors}). Another 10.6\% of errors were caused by either the subject or object entity being completely misaligned with the annotation. We identify some of these as co-reference errors, such as the Subject example in Table \ref{tab:errors}. Evaluation for RE systems often ignores other mentions of an entity. We believe co-reference resolution has not been properly explored within RE evaluation and this may open interesting opportunities for future work.

Finally, it is worth noting that only 1.8\% of errors were due to the wrong relation type being predicted between entities. We consider this to be a strong indicator of the quality of annotated relations between entities. However, we also observed that 72.2\% of the errors were caused by incorrect predictions or missing annotations, highlighting the main shortcoming of our annotation procedure. Our approach is based on annotated hyperlinks in Wikipedia and relations in Wikidata, which can result in recall issues where entities in the text are not identified as hyperlinks or relational facts are not present in Wikidata.

\section{Conclusions}
In this paper, we have addressed some of the key issues facing current multilingual relation extraction datasets by presenting two new resources: SRED\textsuperscript{FM} and RED\textsuperscript{FM}. SRED\textsuperscript{FM} is an automatically annotated dataset covering 18 languages, 400 relation types, 13 entity types, and more than 40 million triplet instances, while RED\textsuperscript{FM} is a smaller, humanly-revised dataset for seven languages. 
We improved the quality of the entity type annotations in these datasets by using a Transformer-based NER classifier. We also introduced the Triplet Critic, a cross-encoder that is trained on annotated data to predict whether a given context entails a triplet. We demonstrated the utility of these new resources by training new, capable multilingual relation extraction models and evaluating them using our supervised data. We also presented mREBEL, the first multilingual end-to-end relation extraction system that extracts triplets, including entity types. Our work thus contributes to the development of better multilingual relation extraction systems and provides valuable resources for future research. 

\section{Limitations}
There are several limitations to the work presented in this paper that need to be acknowledged.

First, the SRED\textsuperscript{FM} and RED\textsuperscript{FM} datasets are based on Wikipedia and Wikidata, which means they may not cover all possible relation types or entities. In addition, the quality of the annotations in these datasets may be influenced by the biases and limitations of these sources.

Second, the Triplet Critic is trained on a small subset of the SRED\textsuperscript{FM} dataset, which may limit its ability to generalize to other relation types or languages. Additionally, the performance of the Triplet Critic may be affected by the quality of the annotations used to train it.

Third, the authors of this work are native speakers of some of the languages tackled in this work and external native speakers created the annotation guidelines. However, for some of the automatically-annotated languages, there were no native speakers involved. Additionally, the qualitative error analysis does not include Arabic or Chinese examples, as neither of the authors of the paper is proficient in those languages.

Finally, the mREBEL system is based on a Transformer architecture, which may not be optimal for all relation extraction tasks. It is possible that other types of model, such as graph neural networks or rule-based systems, could outperform mREBEL on certain relation types or languages.

Overall, the results presented in this paper should be interpreted in the context of these limitations. Further research is needed to address these limitations and to improve the performance of multilingual relation extraction systems.

\section{Ethics Statement}
In this work, we present two new relation extraction datasets, RED\textsuperscript{FM} and SRED\textsuperscript{FM}, which are created using distant supervision techniques and the use of human annotation to filter out false positives. We believe that our datasets will help advance the field of relation extraction by providing a high-quality multilingual resource for researchers and practitioners.

We take the ethical considerations of our work seriously. The annotation of the RED\textsuperscript{FM} dataset is based on existing triplets in Wikidata, which may not always reflect the true relation between entities in a given text. Moreover, the use of human annotation ensures a higher level of accuracy in our dataset, but it also raises ethical considerations. We recognize that human annotation may contain errors or biases. Therefore, we encourage researchers to use our dataset with caution and to perform thorough evaluations of their methods. Additionally, we are transparent about our annotation costs and payment to human annotators.

In conclusion, we believe that our dataset and the research it enables will contribute positively to the field of relation extraction, but we also acknowledge that there are ethical considerations that need to be taken into account when using it.

\section*{Acknowledgments}
\begin{center}
\noindent
\begin{minipage}{0.1\linewidth}
    \raisebox{-0.25\height}{\includegraphics[trim =0mm 5mm 5mm 5mm,clip,scale=0.045]{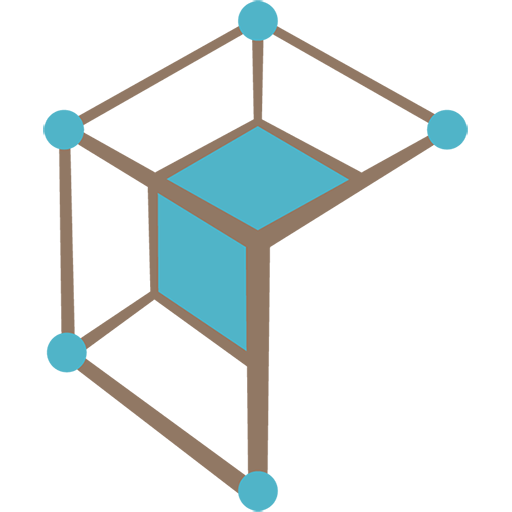}}

\end{minipage}
\hspace{0.005\linewidth}
\begin{minipage}{0.72\linewidth}
The authors gratefully acknowledge the support of the European Union’s Horizon 2020 research project \textit{Knowledge Graphs at Scale} (KnowGraphs) under the Marie Skłodowska-Curie grant agreement No. 860801.

  \vspace{1ex}
\end{minipage}
\hspace{0.005\linewidth}
\begin{minipage}{0.1\linewidth}
  \vspace{0.05cm}
\raisebox{-0.25\height}{\includegraphics[trim =0mm 5mm 5mm 5mm,clip,scale=0.060]{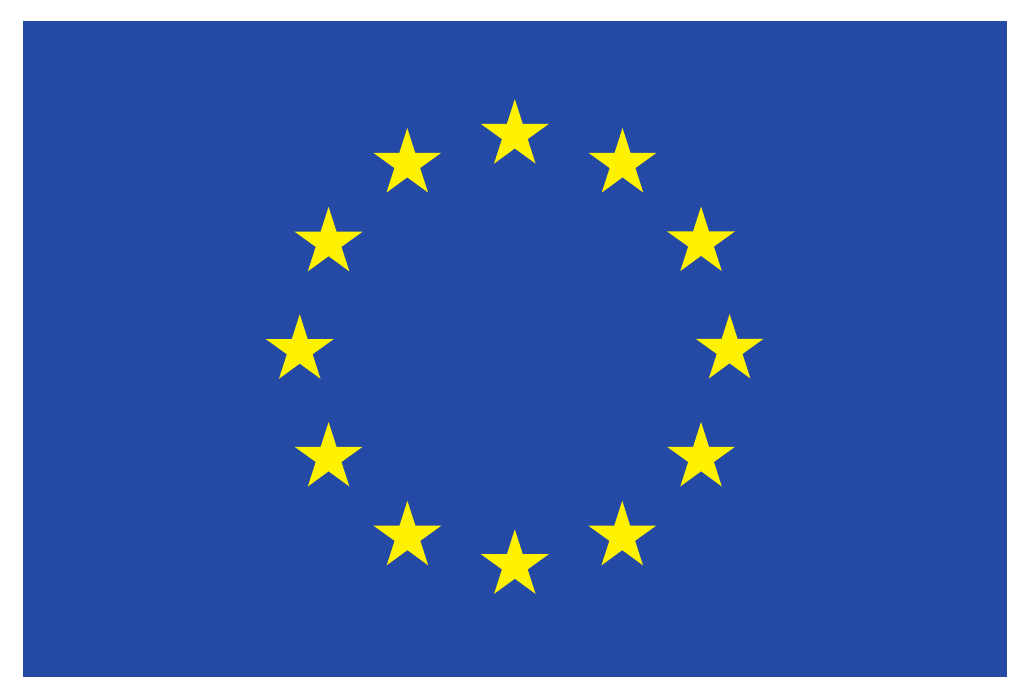}}
  \vspace{0.05cm}
\end{minipage}\\
\end{center}

This research has been carried out while Pere-Lluís Huguet Cabot and Simone Tedeschi were enrolled in the Italian National Doctorate on Artificial Intelligence run by Sapienza University of Rome and with the support of the PNRR MUR project PE0000013-FAIR.

We sincerely thank all annotators who took part in the task as this work would not have been possible without their contribution.
\bibliography{anthology,custom}
\bibliographystyle{acl_natbib}

\appendix

\section{cRocoDiLe}
\label{sec:crocodile}

CRocoDiLe \cite{huguet-cabot-navigli-2021-rebel-relation}, based on \citet{elsahar-etal-2018-rex}, extracts relational information in Wikipedia abstracts, i.e., the text before the Table of contents. It links the entities present in the text as hyperlinks, together with dates and values, to Wikidata entities using \texttt{wikimapper}\footnote{\url{https://pypi.org/project/wikimapper/}}. The original implementation is compatible with Wikipedia dumps in any language; however, its dates and numbers linker were English-specific. We use regex to extract dates and values in all the languages this work covers. 

In the original work, they filtered triplets using NLI. For each triplet, they input the text containing both entities from the Wikipedia abstract, and the triplet in their surface forms, \texttt{subject} + \texttt{relation} + \texttt{object}, separated by the \texttt{<sep>} token. If the score was less than 0.75 for the entailment class, it was removed to ensure higher precision. In our work, we set a lower threshold, 0.1, since we further filter triplets using manual annotation or our Critic model.

\section{Annotation}
\label{sec:appendix_annotation}
We employ Mechanical Turk for annotation purposes. Each annotator was paid 0.1\$ for every ten instances annotated, constituting 1 HIT, an average of \$10 hourly rate. We restrict annotators to countries where each of the languages is spoken, plus the USA. We manually screen annotators in each language separately by having them annotate a small sample of fewer than 10 HITs, and allowing only those who correctly performed the task to annotate the final corpus. 

\begin{figure*}
    \centering
    \includegraphics[width=\textwidth]{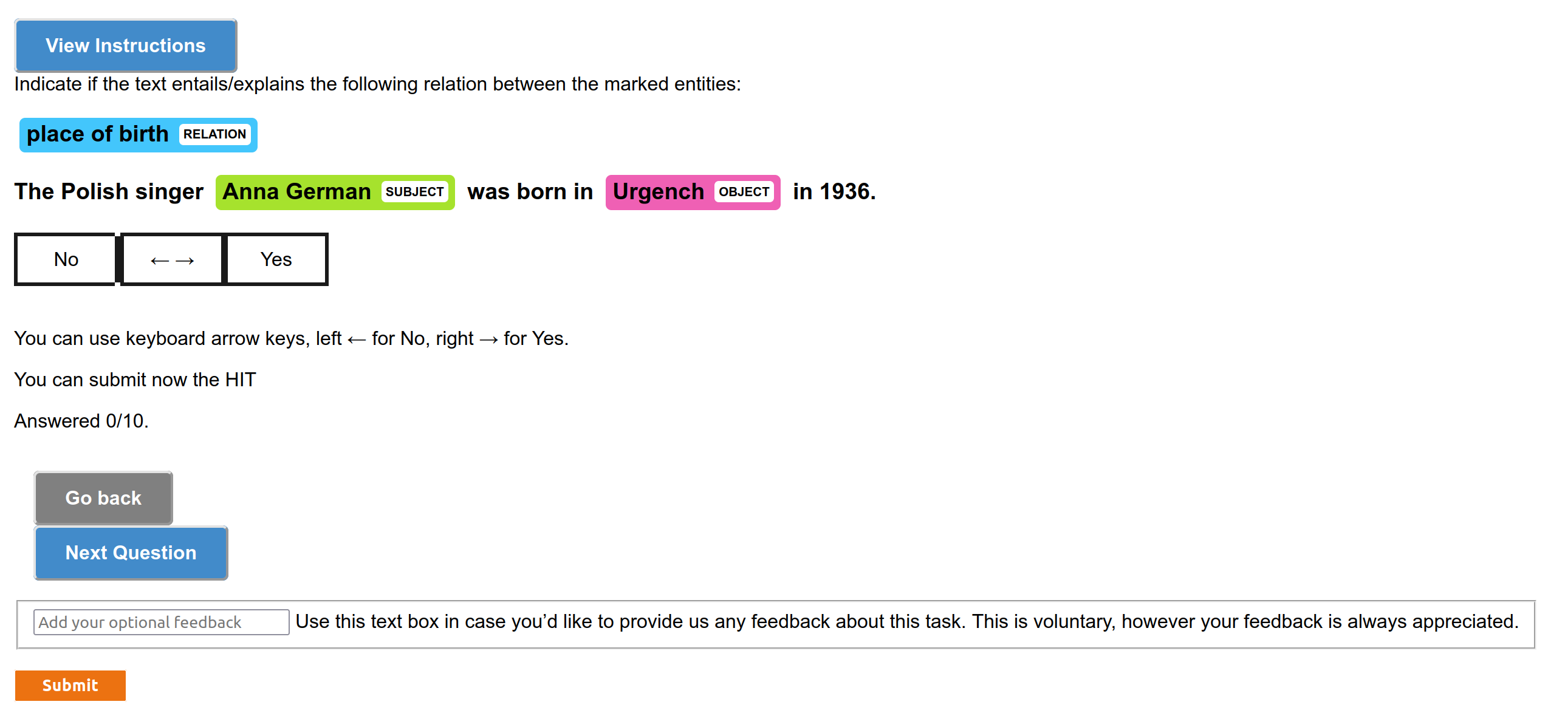}    
    \caption{Annotation example from the MT platform. The interface was translated to each language by native speakers.} 
    \label{fig:annotation}
\end{figure*}

Annotators were presented descriptions for each relation, which they could check at any time by hovering the label or opening the instructions. The English descriptions are:
\begin{itemize}
    \item \textbf{located in the administrative territorial entity}: the item is located on the territory of the following administrative entity
    \item \textbf{country:} sovereign state of this item (not to be used for human beings)
    \item \textbf{instance of:} that class of which this subject is a particular example and member
    \item \textbf{shares border with:} countries or administrative subdivisions, of equal level, that this item borders, either by land or water
    \item \textbf{part of:} object of which the subject is a part
    \item \textbf{capital:} seat of government of a country, province, state or other type of administrative territorial entity
    \item \textbf{follows:} immediately prior item in a series of which the subject is a part
    \item \textbf{headquarters location:} city, where an organization's headquarters is or has been situated
    \item \textbf{located in or next to body of water:} sea, lake, river or stream
    \item \textbf{sport:} sport that the subject participates or participated in or is associated with
    \item \textbf{subsidiary:} subsidiary of a company or organization; generally a fully owned separate corporation
    \item \textbf{member of:} organization, club or musical group to which the subject belongs
    \item \textbf{owned by:} owner of the subject
    \item \textbf{manufacturer:} manufacturer or producer of this product
    \item \textbf{genre:} creative work's genre or an artist's field of work (P101)
    \item \textbf{located on terrain feature:} located on the specified landform
    \item \textbf{child:} subject has object as child
    \item \textbf{author:} main creator(s) of a written work (use on works, not humans); use P2093 when Wikidata item is unknown or does not exist
    \item \textbf{named after:} entity or event that inspired the subject's name, or namesake (in at least one language)
    \item \textbf{country of origin:} country of origin of this item (creative work, food, phrase, product, etc.)
    \item \textbf{replaces:} person, state or item replaced
    \item \textbf{inception:} date or point in time when the subject came into existence as defined
    \item \textbf{cast member:} actor in the subject production
    \item \textbf{subclass of:} next higher class or type; all instances of these items are instances of those items; this item is a class (subset) of that item
    \item \textbf{league:} league in which team or player plays or has played in
    \item \textbf{developer:} organization or person that developed the item
    \item \textbf{location:} location of the object, structure or event
    \item \textbf{occupation:} occupation of a person
    \item \textbf{spouse:} the subject has the object as their spouse (husband, wife, partner, etc.)
    \item \textbf{characters:} characters which appear in this item (like plays, operas, operettas, books, comics, films, TV series, video games)
    \item \textbf{notable work:} notable scientific, artistic or literary work, or other work of significance among subject's works
    \item \textbf{place of birth:} most specific known (e.g. city instead of country, or hospital instead of city) birth location of a person, animal or fictional character
    \item \textbf{mouth of the watercourse:} the body of water to which the watercourse drains
    \item \textbf{country of citizenship:} the object is a country that recognizes the subject as its citizen
    \item \textbf{founded by:} founder or co-founder of this organization, religion or place
    \item \textbf{director:} director(s) of film, TV-series, stageplay, video game or similar
    \item \textbf{sibling:} the subject and the object have the same parents (brother, sister, etc.)
    \item \textbf{participant:} person, group of people or organization (object) that actively takes/took part in an event or process (subject) 
\end{itemize} 
Figure \ref{fig:annotation} shows the annotation interface provided to the annotators.

\section{Results} \label{app:extra_results}
\begin{figure}[!tb]
    \centering
    \includegraphics[width=\columnwidth]{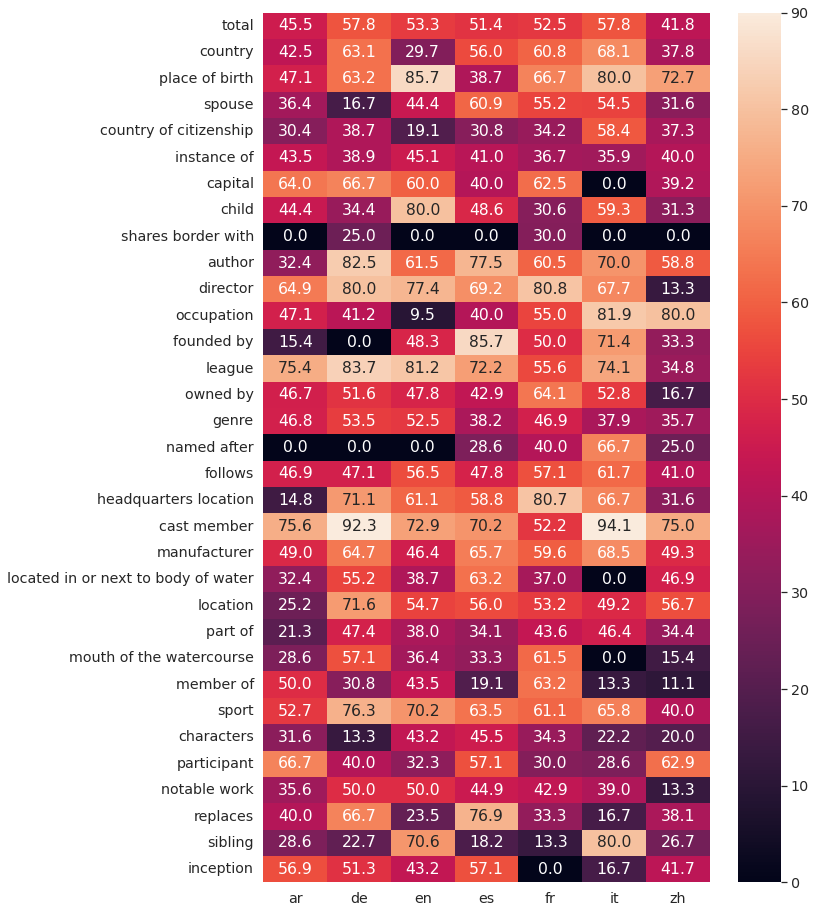} 
    \caption{Results for mREBEL$^{T}_{32}$ on RED\textsuperscript{FM} without fine-tuning.} 
    \label{fig:results_redfm_nofine}
\end{figure}

\begin{figure}[!tb]
    \centering
    \includegraphics[width=\columnwidth]{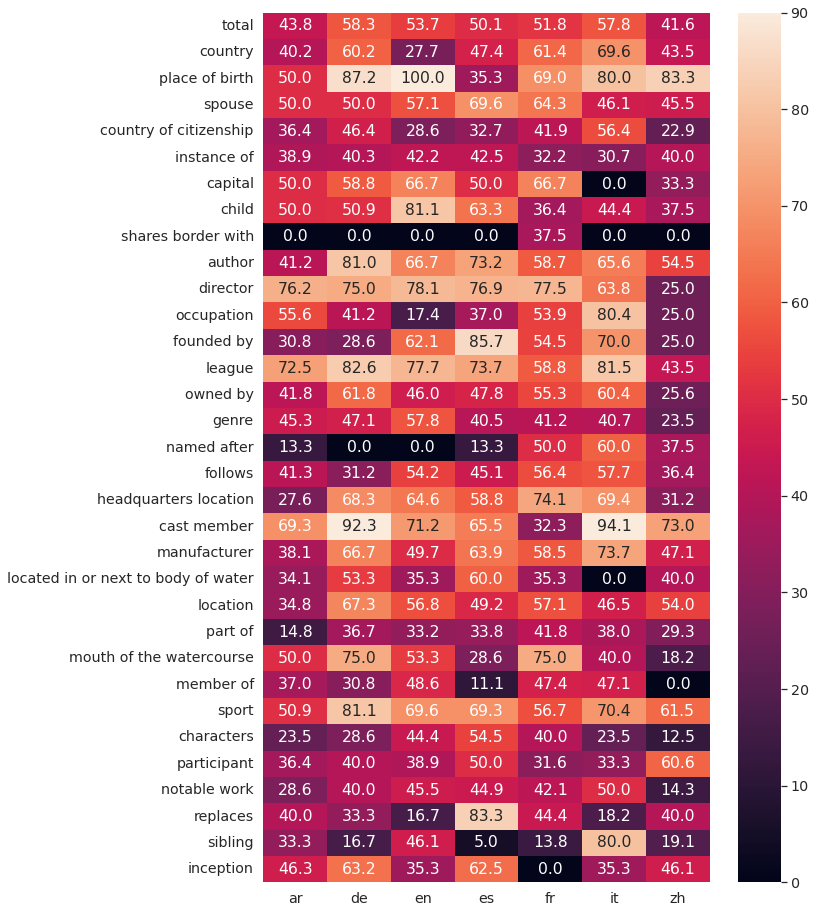} 
\caption{Results for mREBEL$^{T}_{32}$ on RED\textsuperscript{FM} with fine-tuning.}     \label{fig:results_redfm_fine}
\end{figure}

In this Section, we provide more results concerning our mREBEL model. Specifically, in Figure \ref{fig:results_redfm_nofine} we provide a heatmap that shows the scores attained by mREBEL$^{T}_{32}$ (without fine-tuning) on each of the 32 relations covered by RED\textsuperscript{FM}, and for each of its 7 languages. Similarly, in Figure \ref{fig:results_redfm_fine}, we report the scores obtained by the fine-tuned version of mREBEL$^{T}_{32}$. By looking at these two heatmaps, it is easy to identify our model's strengths and weaknesses across relations and languages. We can see how relations such as \textit{named after} or \textit{shares border with} had low scores, probably due to their lower frequency at evaluation time, where a few errors lead to a low score. On the other hand, domain-specific relations such as \textit{cast member}, \textit{league} or \textit{author} show a strong performance on most languages.

\begin{landscape}
\begin{table}[t]
\centering
\resizebox{1.5\textwidth}{!}{%
\begin{tabular}{l|rrrrr|r||rrrrrrr|r||rrrrrrr|r}\toprule\toprule
                                             & \multicolumn{6}{c}{\textbf{Train}}                                                                                                                                                                           & \multicolumn{8}{c}{\textbf{Validation}}                                                                                                                                                                                                                                          & \multicolumn{8}{c}{\textbf{Test}}                                                                                                                                                                                                                                                \\
                                             & \multicolumn{1}{c}{\textbf{de}} & \multicolumn{1}{c}{\textbf{en}} & \multicolumn{1}{c}{\textbf{es}} & \multicolumn{1}{c}{\textbf{fr}} & \multicolumn{1}{c}{\textbf{it}} & \multicolumn{1}{c}{\textbf{Total}} & \multicolumn{1}{c}{\textbf{ar}} & \multicolumn{1}{c}{\textbf{de}} & \multicolumn{1}{c}{\textbf{en}} & \multicolumn{1}{c}{\textbf{es}} & \multicolumn{1}{c}{\textbf{fr}} & \multicolumn{1}{c}{\textbf{it}} & \multicolumn{1}{c}{\textbf{zh}} & \multicolumn{1}{c}{\textbf{Total}} & \multicolumn{1}{c}{\textbf{ar}} & \multicolumn{1}{c}{\textbf{de}} & \multicolumn{1}{c}{\textbf{en}} & \multicolumn{1}{c}{\textbf{es}} & \multicolumn{1}{c}{\textbf{fr}} & \multicolumn{1}{c}{\textbf{it}} & \multicolumn{1}{c}{\textbf{zh}} & \multicolumn{1}{c}{\textbf{Total}} \\ \midrule
\textbf{location}                            & 896                             & 1071                            & 542                             & 602                             & 528                             & 3639                               & 57                              & 53                              & 99                              & 57                              & 54                              & 55                              & 77                              & 452                                & 73                              & 55                              & 76                              & 61                              & 79                              & 62                              & 100                             & 506                                \\
\textbf{instance of}                         & 639                             & 1025                            & 511                             & 603                             & 589                             & 3367                               & 109                             & 74                              & 116                             & 57                              & 78                              & 105                             & 48                              & 587                                & 80                              & 66                              & 116                             & 64                              & 86                              & 115                             & 41                              & 568                                \\
\textbf{country}                             & 865                             & 535                             & 646                             & 703                             & 548                             & 3297                               & 136                             & 105                             & 70                              & 79                              & 119                             & 166                             & 104                             & 779                                & 105                             & 127                             & 42                              & 104                             & 142                             & 167                             & 92                              & 779                                \\
\textbf{part of}                             & 301                             & 668                             & 353                             & 393                             & 248                             & 1963                               & 35                              & 29                              & 61                              & 39                              & 57                              & 34                              & 42                              & 297                                & 23                              & 23                              & 88                              & 31                              & 29                              & 30                              & 39                              & 263                                \\
\textbf{follows}                             & 166                             & 301                             & 210                             & 234                             & 294                             & 1205                               & 52                              & 34                              & 47                              & 27                              & 58                              & 48                              & 35                              & 301                                & 36                              & 16                              & 47                              & 37                              & 35                              & 54                              & 21                              & 246                                \\
\textbf{manufacturer}                        & 258                             & 307                             & 218                             & 201                             & 207                             & 1191                               & 92                              & 73                              & 59                              & 42                              & 72                              & 49                              & 61                              & 448                                & 51                              & 51                              & 77                              & 35                              & 74                              & 49                              & 30                              & 367                                \\
\textbf{sport}                               & 203                             & 288                             & 172                             & 268                             & 249                             & 1180                               & 45                              & 51                              & 101                             & 41                              & 75                              & 103                             & 11                              & 427                                & 43                              & 39                              & 131                             & 39                              & 61                              & 100                             & 6                               & 419                                \\
\textbf{owned by}                            & 199                             & 398                             & 170                             & 198                             & 159                             & 1124                               & 38                              & 28                              & 55                              & 24                              & 33                              & 31                              & 30                              & 239                                & 33                              & 30                              & 49                              & 22                              & 37                              & 28                              & 21                              & 220                                \\
\textbf{located in or next to body of water} & 169                             & 326                             & 252                             & 159                             & 100                             & 1006                               & 17                              & 4                               & 24                              & 7                               & 10                              & 13                              & 23                              & 98                                 & 23                              & 15                              & 16                              & 10                              & 12                              & 2                               & 26                              & 104                                \\
\textbf{member of}                           & 125                             & 347                             & 163                             & 159                             & 129                             & 923                                & 19                              & 11                              & 20                              & 16                              & 13                              & 7                               & 19                              & 105                                & 15                              & 20                              & 49                              & 18                              & 23                              & 12                              & 11                              & 148                                \\
\textbf{genre}                               & 183                             & 272                             & 145                             & 187                             & 128                             & 915                                & 23                              & 33                              & 59                              & 29                              & 34                              & 72                              & 14                              & 264                                & 28                              & 34                              & 67                              & 44                              & 45                              & 68                              & 11                              & 297                                \\
\textbf{author}                              & 150                             & 209                             & 120                             & 139                             & 127                             & 745                                & 29                              & 27                              & 51                              & 23                              & 67                              & 48                              & 28                              & 273                                & 35                              & 58                              & 46                              & 41                              & 44                              & 29                              & 14                              & 267                                \\
\textbf{headquarters location}               & 146                             & 226                             & 110                             & 137                             & 98                              & 717                                & 20                              & 28                              & 28                              & 11                              & 13                              & 17                              & 15                              & 132                                & 9                               & 19                              & 35                              & 9                               & 26                              & 21                              & 17                              & 136                                \\
\textbf{capital}                             & 112                             & 191                             & 105                             & 109                             & 104                             & 621                                & 7                               & 9                               & 7                               & 4                               & 9                               & 2                               & 12                              & 50                                 & 12                              & 8                               & 4                               & 2                               & 6                               & 3                               & 14                              & 49                                 \\
\textbf{child}                               & 107                             & 147                             & 104                             & 134                             & 102                             & 594                                & 23                              & 12                              & 16                              & 9                               & 14                              & 12                              & 25                              & 111                                & 15                              & 23                              & 15                              & 28                              & 36                              & 15                              & 33                              & 165                                \\
\textbf{notable work}                        & 86                              & 204                             & 61                              & 137                             & 97                              & 585                                & 21                              & 14                              & 47                              & 10                              & 37                              & 34                              & 19                              & 182                                & 19                              & 20                              & 38                              & 17                              & 31                              & 18                              & 4                               & 147                                \\
\textbf{mouth of the watercourse}            & 98                              & 156                             & 205                             & 47                              & 72                              & 578                                & 2                               & 1                               & 15                              & 4                               & 2                               & 6                               & 3                               & 33                                 & 4                               & 4                               & 6                               & 3                               & 6                               & 2                               & 3                               & 28                                 \\
\textbf{inception}                           & 65                              & 335                             & 90                              & 13                              & 69                              & 572                                & 55                              & 10                              & 46                              & 10                              & 3                               & 5                               & 21                              & 150                                & 48                              & 21                              & 36                              & 9                               & 0                               & 9                               & 13                              & 136                                \\
\textbf{named after}                         & 139                             & 192                             & 81                              & 80                              & 66                              & 558                                & 10                              & 11                              & 6                               & 4                               & 5                               & 5                               & 7                               & 48                                 & 8                               & 6                               & 3                               & 9                               & 10                              & 6                               & 9                               & 51                                 \\
\textbf{occupation}                          & 172                             & 40                              & 105                             & 139                             & 70                              & 526                                & 34                              & 12                              & 12                              & 13                              & 49                              & 89                              & 12                              & 221                                & 17                              & 18                              & 11                              & 14                              & 43                              & 87                              & 4                               & 194                                \\
\textbf{shares border with}                  & 77                              & 56                              & 153                             & 117                             & 63                              & 466                                & 3                               & 2                               & 3                               & 1                               & 4                               & 1                               & 3                               & 17                                 & 6                               & 4                               & 4                               & 1                               & 7                               & 1                               & 5                               & 28                                 \\
\textbf{participant}                         & 48                              & 192                             & 86                              & 66                              & 51                              & 443                                & 6                               & 0                               & 16                              & 5                               & 8                               & 17                              & 6                               & 58                                 & 7                               & 2                               & 20                              & 4                               & 13                              & 4                               & 14                              & 64                                 \\
\textbf{country of citizenship}              & 111                             & 90                              & 105                             & 71                              & 64                              & 441                                & 23                              & 19                              & 13                              & 18                              & 20                              & 111                             & 29                              & 233                                & 19                              & 26                              & 7                               & 21                              & 18                              & 108                             & 44                              & 243                                \\
\textbf{founded by}                          & 81                              & 134                             & 58                              & 97                              & 68                              & 438                                & 16                              & 13                              & 16                              & 2                               & 10                              & 6                               & 7                               & 70                                 & 9                               & 4                               & 17                              & 7                               & 10                              & 8                               & 3                               & 58                                 \\
\textbf{place of birth}                      & 128                             & 94                              & 90                              & 101                             & 23                              & 436                                & 14                              & 18                              & 7                               & 12                              & 12                              & 1                               & 4                               & 68                                 & 8                               & 20                              & 3                               & 10                              & 11                              & 2                               & 6                               & 60                                 \\
\textbf{replaces}                            & 78                              & 118                             & 55                              & 70                              & 70                              & 391                                & 5                               & 4                               & 12                              & 1                               & 6                               & 7                               & 16                              & 51                                 & 3                               & 9                               & 11                              & 7                               & 6                               & 7                               & 11                              & 54                                 \\
\textbf{cast member}                         & 53                              & 145                             & 95                              & 12                              & 43                              & 348                                & 35                              & 38                              & 73                              & 20                              & 20                              & 18                              & 4                               & 208                                & 60                              & 24                              & 102                             & 26                              & 12                              & 16                              & 29                              & 269                                \\
\textbf{league}                              & 81                              & 125                             & 42                              & 38                              & 47                              & 333                                & 24                              & 24                              & 33                              & 18                              & 8                               & 24                              & 20                              & 151                                & 35                              & 22                              & 49                              & 18                              & 8                               & 11                              & 12                              & 155                                \\
\textbf{characters}                          & 39                              & 113                             & 38                              & 64                              & 46                              & 300                                & 15                              & 9                               & 19                              & 15                              & 16                              & 9                               & 4                               & 87                                 & 8                               & 11                              & 24                              & 10                              & 13                              & 13                              & 9                               & 88                                 \\
\textbf{director}                            & 66                              & 69                              & 37                              & 66                              & 58                              & 296                                & 11                              & 22                              & 16                              & 6                               & 35                              & 26                              & 4                               & 120                                & 20                              & 19                              & 30                              & 12                              & 22                              & 32                              & 7                               & 142                                \\
\textbf{spouse}                              & 32                              & 76                              & 49                              & 57                              & 41                              & 255                                & 3                               & 8                               & 8                               & 6                               & 13                              & 7                               & 6                               & 51                                 & 8                               & 5                               & 9                               & 11                              & 17                              & 5                               & 9                               & 64                                 \\
\textbf{sibling}                             & 36                              & 54                              & 23                              & 51                              & 39                              & 203                                & 3                               & 1                               & 5                               & 1                               & 2                               & 1                               & 12                              & 25                                 & 4                               & 12                              & 7                               & 9                               & 13                              & 2                               & 5                               & 52                                 \\ \midrule
\textbf{total}                               & 5909                            & 8504                            & 5194                            & 5452                            & 4597                            & 29656                              & 982                             & 777                             & 1160                            & 611                             & 956                             & 1129                            & 721                             & 6336                               & 864                             & 811                             & 1235                            & 733                             & 975                             & 1086                            & 663                             & 6367                              \\\bottomrule\bottomrule
\end{tabular}%

}
\caption{Breakdown for RED\textsuperscript{FM}.}
\label{tab:REDFM}
\end{table}
\end{landscape}

\section{Reproducibility}
\label{app:repro}
Experiments were performed using a single NVIDIA 3090 GPU with 64GB of RAM and Intel\textsuperscript{\textregistered} Core\textsuperscript{\texttrademark} i9-10900KF CPU.

The hyperparameters were manually tuned on the validation sets for each dataset, but mostly left at default values for mBART. The ones used for the final results can be found in Table \ref{tab:params}.

\begin{table}[t!]
\centering
\resizebox{0.5\textwidth}{!}{
\begin{tabular}{lccccc}
\hline
                   & \textbf{Learning Rate} & \textbf{Warm-up} & \textbf{Batch size} & \multicolumn{1}{l}{\textbf{Max Steps}} \\ \hline
\textbf{mREBEL$_{400}$}                    & $10^{-5}  $                 & 5000 steps                       & 32                  & 1.6M                                     \\
\textbf{mREBEL$_{32}$}                    & $10^{-5}  $                 & 5000 steps                      & 32                  & + 264K                                     \\ 
\textbf{mREBEL$_{B400}$}                    & $5 \times 10^{-5}  $                 & 5000 steps                      & 32                  & 1M                                     \\ \midrule

\textbf{SMILER}                 & $5 \times 10^{-5}  $                 & 3000 steps                            & 32                  & + 10K                                     \\
\textbf{RED\textsuperscript{FM}}                    & $10^{-5}  $                 & 1000 steps                        & 32                  & + 10K                                      \\ \hline
\end{tabular}}
\caption{Hyperparameters for the different datasets. Top half shows used the values used in the pretraining phase, while the bottom part shows those used during fine-tuning.}
\label{tab:params}
\end{table}

\section{Data}
In Table \ref{tab:REDFM}, we provide data statistics for our RED\textsuperscript{FM} dataset. Specifically, for each of the 7 languages, we report the number of instances for each relation in the corresponding training, validation and test sets.
\label{sec:appendix_data}

\end{document}